\documentclass[11pt,a4paper]{article}
\usepackage[hyperref]{emnlp2018}
\usepackage{times}
\usepackage{latexsym}
\usepackage{multirow}
\usepackage{booktabs}
\usepackage{linguex}
\usepackage{array}
\usepackage{url}
\usepackage[colorinlistoftodos,prependcaption,textsize=scriptsize]{todonotes}

\newcommand{\citeNP}[1]{\citeauthor{#1} \citeyear{#1}}

\aclfinalcopy 
\def\aclpaperid{1799} 

\title{Targeted Syntactic Evaluation of Language Models}

\author{Rebecca Marvin \\
    Department of Computer Science\\
  Johns Hopkins University \\
  {\tt becky@jhu.edu} \\\And
  Tal Linzen \\
  Department of Cognitive Science\\
  Johns Hopkins University \\
  {\tt tal.linzen@jhu.edu} \\}

\date{}

\begin{document}
\maketitle
\begin{abstract}
    We present a dataset for evaluating the grammaticality of the predictions of a language model. We automatically construct a large number of minimally different pairs of English sentences, each consisting of a grammatical and an ungrammatical sentence. The sentence pairs represent different variations of structure-sensitive phenomena: subject-verb agreement, reflexive anaphora and negative polarity items. We expect a language model to assign a higher probability to the grammatical sentence than the ungrammatical one. In an experiment using this data set, an LSTM language model performed poorly on many of the constructions. Multi-task training with a syntactic objective (CCG supertagging) improved the LSTM's accuracy, but a large gap remained between its performance and the accuracy of human participants recruited online. This suggests that there is considerable room for improvement over LSTMs in capturing syntax in a language model.
\end{abstract}

\setlength{\Exlabelwidth}{0.25em}
\setlength{\SubExleftmargin}{1.35em}

\section{Introduction}

A language model (LM) defines a probability distribution over sequences of words. Recent technological advances have led to an explosion of neural network-based LM architectures. The most popular ones are based on recurrent neural networks (RNNs) \cite{elman1990finding,mikolov2010recurrent}, in particular Long Short-Term Memory networks (LSTMs) \cite{hochreiter1997long}. While a large number of alternative architectures have been proposed in the past few years, LSTMs are still highly competitive \cite{melis2018state}.

Language models are typically evaluated using perplexity: it is considered desirable for an LM to assign a high probability to held-out data from the same corpus as the training data. This measure conflates multiple sources of success (or failure) in predicting the next word: common collocations, semantics, pragmatics, syntax, and so on. The quality of the \textbf{syntactic} predictions made by the LM is arguably particularly difficult to measure using perplexity: since most sentences are grammatically simple and most words can be predicted from their local context, perplexity rewards LMs primarily for collocational and semantic predictions. 

We propose to supplement perplexity with a metric that assesses whether the probability distribution defined by the model conforms to the grammar of the language. Following previous work \cite{lau2017grammaticality,linzen2016assessing,Gulordava18}, we suggest that given two sentences that differ minimally from each other, one of which is grammatical and the other which is not, it is desirable for the model to assign a higher probability to the grammatical one. 

The value of this approach can be illustrated with a recent study by \newcite{tran2018importance}, where a standard LSTM language model was compared to an attention-only LM without recurrence \cite{vaswani2017attention}. Although the attention-only model had somewhat better perplexity on the validation set, when the models were tested specifically on challenging subject-verb agreement dependencies, the attention-only model made three times as many errors as the LSTM. In other words, the LSTM learned more robust syntactic representations, but this advantage was not reflected in its average perplexity on the corpus, since syntactically challenging sentences are relatively infrequent.

    Previous work on targeted syntactic evaluation of language models has identified syntactically challenging sentences in corpora \cite{linzen2016assessing,Gulordava18}. While evaluation on naturally occurring examples is appealing, this approach has its limitations (see Section \ref{sec:approach}). In particular, syntactically challenging examples are sparsely represented in a corpus, their identification requires a clean parsed corpus, and naturally occurring sentences are difficult to control for confounds. We contrast the naturalistic approach with a constructed dataset, which allows us to examine a much larger range of specific grammatical phenomena than has been possible before. We use templates to automatically create our test sentences, making it possible to generate a large test set while maintaining experimental control over our materials as well as a balanced number of examples of each phenomenon.

    We test three LMs on the data set we develop: an \emph{n}-gram baseline, an RNN LM trained on an unannotated corpus, and an RNN LM trained on a multitask objective: language modeling and Combinatory Categorial Grammar (CCG) supertagging \cite{bangalore1999supertagging}. We also conduct a human experiment using the same materials. The \mbox{\emph{n}-gram} baseline largely performed at chance, suggesting that good performance on the task requires syntactic representations. The RNN LMs performed well on simple cases, but struggled on more complex ones. Multi-task training with a supervised syntactic objective improved the performance of the RNN, but it was still much weaker than humans. This suggests that our data set is challenging, especially when explicit syntactic supervision is not available, and can therefore motivate richer language modeling architectures.

\section{Overview of the approach}\label{sec:approach}

\subsection{Grammaticality and LM probability}

How should grammaticality be captured in the probability distribution defined by an LM? The most extreme position would be that a language model should assign a probability of zero to ungrammatical sentences. For most applications, some degree of error tolerance is desirable, and it is not practical to assign a sentence a probability of exactly zero.\footnote{Nor is it possible to have a threshold $\epsilon$ such that all grammatical sentences have probability higher than $\epsilon$ and all ungrammatical sentences have probability lower than $\epsilon$, for the simple reason that there is an infinite number of grammatical sentences \cite{lau2017grammaticality}.} Following \citet{linzen2016assessing} and \citet{Gulordava18}, our desideratum for the language model is more modest: if two closely matched sentence differ only in their grammaticality, the probability of the grammatical sentence should be higher than the probability of the ungrammatical one. For example, the following minimal pair illustrates the fact that third-person present English verbs agree with the number of their subject:

\ex.\textit{Simple agreement}:\label{ex:simple_agreement}
    \a.The author \underline{laughs}.\label{ex:simple_agreement_grammatical}
    \b.*The author \underline{laugh}.\label{ex:simple_agreement_ungrammatical}

   We expect the probability of \ref{ex:simple_agreement_grammatical} to be higher than the probability of \ref{ex:simple_agreement_ungrammatical}. Previous work has simplified this setting further by comparing the probability that the LM assigns to a single word that is the locus of ungrammaticality. In \ref{ex:simple_agreement}, for example, the LM would be fed the first two words of the sentence, and would be considered successful on the task if it predicts $P(\textit{laughs}) > P(\textit{laugh})$. 

   The prediction setting is only applicable when the locus of ungrammaticality is a single word, rather than, say, the interaction between two words; moreover, the information needed to make the grammaticality decision needs to be available in the left context of the locus of grammaticality. These conditions do not always hold. Negative polarity items (NPIs), for example, are words like \textit{any} and \textit{ever} that can only be used in the scope of negation.\footnote{In practice, the conditions that govern the distribution of NPIs are much more complicated, but this first approximation will suffice for the present purposes. For a review, see \newcite{giannakidou2011negative}.} The grammaticality of placing a particular quantifier in the beginning of the sentences in \ref{ex:simple_npi} depends on whether the sentence contains an NPI later on:

   \ex.\textit{Simple NPI:}\label{ex:simple_npi}
    \a.\underline{No} students have ever lived here.
    \b.*\underline{Most} students have ever lived here.

It would not be possible to compare these two sentences using the prediction task. In the current paper, we use the more general setting and compare the probability of the two complete sentences.
    
\subsection{Data set construction}

Previous work has used syntactically complex sentences identified from a parsed corpus. This approach has several limitations. If the corpus is automatically parsed, the risk of a parse error increases with the complexity of the construction \cite{bender2011parser}. If the test set is restricted to sentences with gold parses, it can be difficult or impossible to find a sufficient number of examples of syntactically challenging cases. Moreover, using naturally occurring sentences can introduce confounds that may complicate the interpretation of the experiments \cite{ettinger2018assessing}.

To circumvent these issues, we use templates to automatically construct a large number of English sentence pairs ($\sim$350,000). Our data set includes three phenomena that linguists consider to be sensitive to hierarchical syntactic structure \cite{everaert2015structures,xiang2009illusory}: subject-verb agreement (described in detail in Sections \ref{sec:agreement} and \ref{sec:orc}), reflexive anaphora (Section \ref{sec:reflexive}) and negative polarity items (Section \ref{sec:npi}). 

The templates can be described using non-recursive context-free grammars. We specify the preterminal symbols that make up a syntactic construction and have different terminal symbols that those preterminals could be mapped to. For example, the template for the simple agreement construction illustrated in \ref{ex:simple_agreement} consists of the following rules:

\ex.\a.Simple agreement $\rightarrow$ D\enskip MS\enskip MV
    \b.D $\rightarrow$ \textit{the}
    \c.MS $\rightarrow \{$\textit{author}, \textit{pilot}, $\ldots\}$
    \d.MV $\rightarrow \{$\textit{laughs}, \textit{smiles}, $\ldots\}$
    
We generate all possible combinations of the terminals. The Supplementary Materials provide a full description of all our templates.\footnote{The code, the data set and the Supplementary Materials can be found at \url{https://github.com/BeckyMarvin/LM_syneval}.} 

    While these examples are somewhat artificial, our goal is to isolate the syntactic capabilities of the model; it is in fact beneficial to minimize the semantic or collocational cues that can be used to identify the grammatical sentence. \citeauthor{Gulordava18} took this approach further and constructed ``colorless green ideas'' test cases by substituting random content words into sentences from a corpus. We take a more moderate position and avoid combinations that are very implausible or violate selectional restrictions (e.g., \textit{the apple laughs}). We do this by having separate templates for animate and inanimate subjects and verbs so that the resulting sentences are always reasonably plausible.

\section{Related work}
\label{sec:related_work}

\paragraph{Targeted evaluation:} LM evaluation data sets using challenging prediction tasks have been proposed in the context of semantics and discourse comprehension \cite{Zweig11,paperno2016lambada}. Evaluation sets consisting of challenging syntactic constructions have been constructed for parser evaluation \cite{Rimell09,nivre2010evaluation,bender2011parser}, and minimal pair approaches have been proposed for evaluating image captioning \cite{Shekhar17} and machine translation systems \cite{sennrich2017how}, but no data sets exist that target a range of syntactic constructions for language model evaluation.

\paragraph{Acceptability judgments:} \citet{lau2017grammaticality} compared the ability of different LMs to predict graded human acceptability judgments. The forced-choice approach used in the current paper has been shown to be effective in human acceptability judgment experiments \cite{sprouse2017design}. In some early work, neural networks were trained explicitly to predict acceptability judgments \cite{lawrence1996can,allen1999emergence}; \citet{post2011judging} likewise trained a classifier on top of a parser to predict grammaticality. \citet{warstadt2018neural} use a transfer learning approach, where an unsupervised model is fine-tuned on acceptability prediction. Our work differs from those studies in that we do not advocate providing any explicit grammaticality signal to the LM at any point (``no negative evidence'').

\paragraph{Syntax in LMs:} There have been several proposals over the years to incorporate explicit syntax into LMs to overcome the inability of \emph{n}-gram LMs to model long-distance dependencies \cite{jurafsky1995using,roark2001probabilistic,pauls2012large}. While RNN language models can in principle model longer dependencies \cite{mikolov2010recurrent,linzen2016assessing}, in practice it can still be beneficial to inject syntax into the model. This can be done by combining it with a supervised parser \citep{Dyer16} or other multi-task learning objectives \cite{Enguehard17}. Our work is orthogonal to this area of research, but can be seen as providing a potential opportunity to underscore the advantage of such syntax-infused models.

\section{Data set composition}

This section describes all of the types of sentence pairs included in our data set, which include examples of subject-verb agreement (Sections \ref{sec:agreement} and \ref{sec:orc}), reflexive anaphoras (Section \ref{sec:reflexive}) and negative polarity items (Section \ref{sec:npi}).

\subsection{Subject-verb agreement}
\label{sec:agreement}

Determining the correct number of the verb is trivial in examples such as \ref{ex:simple_agreement} above, in which the sentence only contains a single noun. By contrast, in cases where there are multiple nouns in the sentence, identifying which of them is the subject of a given verb requires understanding the structure of the sentence. In particular, the relevant subject is not necessarily the first noun of the sentence:

\ex.\textit{Agreement in a sentential complement:}
    \a.The bankers knew the officer \underline{smiles}.\label{ex:agr_sc_gram}
    \b.*The bankers knew the officer \underline{smile}.\label{ex:agr_sc_ungram}

    Here the verb \textit{smiles} needs to agree with the embedded subject \textit{officer} rather than the main clause subject \textit{bankers}. The subject is also not necessarily the most recent noun before the verb: when the subject is modified by a phrase, a distracting noun (``attractor'') often intervenes in the linear order of the sentence between the head of the subject and the verb. Two examples of such modifiers are prepositional phrases and relative clauses
 (RCs):

\ex.\textit{Agreement across a prepositional phrase:}\label{ex:agr_pp}
    \a.The farmer near the parents \underline{smiles}.
    \b.*The farmer near the parents \underline{smile}.

\ex.\textit{Agreement across a subject relative clause:}
    \a.The officers that love the skater \underline{smile}.
    \b.*The officers that love the skater \underline{smiles}.

    We include all four possible configurations of noun number for each type of minimal pair; for~\ref{ex:agr_pp}, these would be:\footnote{The slash notation indicates the word that differs between the grammatical and ungrammatical sentence; for example, in \ref{ex:slashnotation}, the full sentence pair would be:

\ex.\a.The farmer near the parent smiles.
    \b.*The farmer near the parent smile.

    }
    
\ex.\a.The farmer near the parent smiles/*smile.\label{ex:slashnotation}
    \b.The farmer near the parents smiles/*smile.
    \c.The farmers near the parent smile/*smiles.
    \d.The farmers near the parents smile/*smiles.
    
Sentences where the two nouns conflict in number are expected to be more challenging, but interpretable errors may certainly occur even when they do not. For example, the model may use the heuristic that sentences with multiple nouns are likely to have a plural verb (a heuristic that would be effective for coordination); alternatively, it might prefer singular verbs to plural ones regardless of whether the subject is singular or plural, simply because the singular form of the verb is more frequent.

Next, in verb phrase (VP) coordination, both of the verbs need to agree with the subject:

\ex.\textit{Short VP coordination:}
\a.The senator smiles and \underline{laughs}.
\b.*The senator smiles and \underline{laugh}.

We had both singular and plural subjects. The number of the verb immediately adjacent to the subject was always grammatical. This problem can in principle be solved with a trigram model (\textit{smiles and laughs} is likely to be a more frequent trigram than \textit{smiles and laugh}); to address this potential concern, we also included a coordination condition with a longer dependency:

\ex.\textit{Long VP coordination:}\vspace{0.2em}\\
The manager writes in a journal every day and likes/*like to watch television shows.

\subsection{Agreement and object relative clauses}
\label{sec:orc}

    We go into greater depth in object relative clauses, which most clearly require a hierarchical representation. In \ref{ex:agr_across_orc} and \ref{ex:agr_in_orc}, the model needs to be able to distinguish the embedded subject (\textit{parents}) from the main clause subject (\textit{farmer}) when making its predictions:

\setlength{\Exlabelwidth}{0.8em}
\setlength{\SubExleftmargin}{1.5em}

\ex.\textit{Agreement across an object relative clause:}\label{ex:agr_across_orc}
    \a.The farmer that the parents love \underline{swims}.
    \b.*The farmer that the parents love \underline{swim}.

\ex.\textit{Agreement in an object relative clause:}\label{ex:agr_in_orc}
    \a.The farmer that the parents \underline{love} swims.
    \b.*The farmer that the parents \underline{loves} swims.

    In keeping with the minimal pair approach, we never introduce two agreement errors at the same time: either the embedded verb or the main verb is incorrectly inflected, but not both.

We include a number of variations on the pattern in \ref{ex:agr_in_orc}. First, we delete the relativizer \textit{that}, with the hypothesis that the absence of an overt cue to structure will make the task more difficult:

\ex.The farmer the parents love/*loves swims.

In another condition, we replace the main subject with an inanimate noun and keep the embedded subject animate. We base this manipulation on human experimental work showing that similar nouns (for example, two animate nouns) are more likely to cause confusion during comprehension than dissimilar nouns, such as an animate and an inanimate noun \cite{vandyke2007interference}:

\ex.The movies that the author likes are/*is good.

For a complete list of all the types of minimal pairs we include, see the Supplementary Materials.

\subsection{Reflexive anaphora}
\label{sec:reflexive}

A reflexive pronoun such as \textit{himself} needs to have an antecedent from which it derives its interpretation. The pronoun needs to agree in number (and gender) with its antecedent:

\ex.\textit{Simple reflexive}:
    \a.The senators embarrassed themselves.\label{ex:simple_refl_gram} 
    \b.*The senators embarrassed herself.\label{ex:simple_refl_ungram} 

    There are structural conditions on the nouns to which a reflexive pronoun can be bound. One of these conditions requires the antecedent to be in the same clause as the reflexive pronoun. For example, \ref{ex:refl_sc_ungram} cannot refer to a context in which \textit{the pilot} embarrassed \textit{the bankers}:

\ex.\textit{Reflexive in a sentential complement}:
    \a.The bankers thought the pilot embarrassed himself.\label{ex:refl_sc_gram}  
    \b.*The bankers thought the pilot embarrassed themselves.\label{ex:refl_sc_ungram} 

Likewise, in the following minimal pair, sentence \ref{ex:refl_rc_ungram} is ungrammatical, because the reflexive pronoun \textit{themselves}, which is part of the main clause, cannot be bound to the noun phrase \textit{the architects}, which is inside an embedded clause:

\ex.\textit{Reflexive across an object relative clause:}
    \a. The manager that the architects like doubted himself.\label{ex:refl_rc_gram}
    \b.*The manager that the architects like doubted themselves.\label{ex:refl_rc_ungram}

\subsection{Negative polarity items}
\label{sec:npi}

Negative polarity items, introduced in example \ref{ex:simple_npi} above, are words that (to a first approximation) need to occur in the context of negation. Crucially for the purposes of the present work, the scope of negation is structurally defined. In particular the negative noun phrase needs to c-command the NPI: the syntactic non-terminal node that dominates the negative noun phrase must also dominate the NPI. This is the case in \ref{ex:npi_gram}, but not in \ref{ex:npi_ungram}, where the negative noun phrase is too deep in the tree to c-command the NPI \textit{ever} \cite{xiang2009illusory,everaert2015structures}.

\ex.\textit{NPI across a relative clause:}\label{ex:npi_across_rc}
    \a.\underline{No} authors that \underline{the} security guards like have ever been famous.\label{ex:npi_gram}
    \b.*\underline{The} authors that \underline{no} security guards like have ever been famous.\label{ex:npi_ungram}

    All of the nouns and verbs in the NPI cases were plural. As in some of the agreement cases, we included a variant of \ref{ex:npi_across_rc} in which the subject was inanimate.

\section{Experimental setup}
\label{sec:lms}

To show how our challenge set can be used to evaluate the syntactic performance of LMs, we trained three LMs with increasing levels of syntactic sophistication. All of the LMs were trained on a 90 million word subset of Wikipedia \cite{Gulordava18}. Our \emph{n}-gram LM and  LSTM LM do not require annotated data. The third model is also an LSTM LM, but it requires syntactically annotated data (CCG supertags).

\paragraph{\emph{N}-gram model:} We trained a 5-gram model on the same 90M word corpus using the SRILM toolkit \citep{stolcke2002srilm} which backs off to smaller \emph{n}-grams using Kneser-Ney smoothing.  

\paragraph{Single-task RNN:} The RNN LM had two layers of 650 LSTM units, a batch size of 128, a dropout rate of 0.2, and a learning rate of 20.0, and was trained for 40 epochs (following the hyperparameters of \citeNP{Gulordava18}).

\paragraph{Multi-task RNN:} In multi-task learning, the system is trained to optimize an objective function that combines the objective functions of several tasks. We combine language modeling with CCG supertagging, a task that predicts for each word in the sentence its CCG supertag \cite{bangalore1999supertagging,lewis2016lstm}. We simply sum the two objective functions with equal weights \cite{Enguehard17}. Early stopping in this model is based on the combined loss on language modeling and supertagging. Supertags provide a large amount of syntactic information about the word; the sequence of supertags of a sentence strongly constrains the possible parses of the sentence. We use supertagging as a ``scaffold'' task \cite{swayamdipta2017frame}: our goal is not to produce a competitive supertagger, but to induce better syntactic representations, which would then lead to improved language modeling. We used CCG-Bank \cite{hockenmaier2007ccgbank} as our CCG corpus.

\newcolumntype{R}[1]{>{\raggedleft\let\newline\\\arraybackslash\hspace{0pt}}m{#1}}
\begin{table*}
\centering
\scalebox{0.85}{
\begin{tabular}{lp{3.5em}p{3.5em}p{3.5em}p{3.5em}R{3.5em}}
\toprule
& RNN & Multitask & \emph{n}-gram & Humans & \# sents\\
\midrule
\textsc{Subject-verb agreement:}\vspace{0.2em}\\
Simple & 0.94 & 1.00 & 0.79 & 0.96 & 280\\
In a sentential complement & 0.99 & 0.93 & 0.79 & 0.93 & 3360\\
Short VP coordination & 0.90 & 0.90 & 0.51 & 0.94 & 1680 \\
Long VP coordination & 0.61 & 0.81 & 0.50 & 0.82 & 800\\
Across a prepositional phrase & 0.57  & 0.69 & 0.50 & 0.85 & 44800\\
Across a subject relative clause  & 0.56  & 0.74  & 0.50 & 0.88 & 22400\\
Across an object relative clause & 0.50 & 0.57 & 0.50 & 0.85 & 44800\\
Across an object relative (no \textit{that})  & 0.52 & 0.52 & 0.50 & 0.82 & 44800\\
In an object relative clause & 0.84 &  0.89 & 0.50 & 0.78 & 44800\\
In an object relative (no \textit{that}) & 0.71 & 0.81 & 0.50 & 0.79 & 44800\\
\midrule
\textsc{Reflexive anaphora:}\vspace{0.2em}\\
Simple & 0.83 & 0.86 & 0.50 & 0.96 & 560\\
In a sentential complement & 0.86 & 0.83 & 0.50 & 0.91 & 6720\\
Across a relative clause & 0.55 & 0.56 & 0.50 & 0.87 & 44800\\
\midrule
\textsc{Negative polarity items:}\vspace{0.2em}\\
Simple & 0.40 & 0.48 & 0.06 & 0.98 & 792\\
Across a relative clause & 0.41 & 0.73 & 0.60 & 0.81 & 31680\\
\bottomrule
\end{tabular}
}
\caption{Overall accuracies for the LSTMs, \emph{n}-gram model and humans on each test case.}
\label{table:overall}
\end{table*}

\paragraph{Human evaluation:}

We designed a human experiment on Amazon Mechanical Turk that mirrored the task that was given to the LMs: both versions of a minimal pair were shown on the screen at the same time, and participants were asked to judge which one of them was more acceptable (for details, see the Supplementary Materials). We emphasize that we do not see human performance on complex syntactic dependencies as setting an upper bound on the performance that we should expect from an LM. There is a rich literature showing that humans make mistakes such as subject-verb agreement errors; in fact, most of the phenomena we test were inspired by work in psycholinguistics that studies these errors \cite{bock1991broken,phillips2011grammatical}. At the same time, while we do not see a reason not to aspire for 100\% accuracy, we are interested in comparing LM and human errors: if the errors are similar, the two systems may be using similar representations.

\section{Results}\label{sec:results}

\paragraph{Local agreement:} The overall accuracy per condition can be seen in Table~\ref{table:overall}. The \emph{n}-gram LM's accuracy was only 79\% for simple agreement and agreement in a sentential complement, both of which can be solved entirely using local context. This is because not all subject and verb combinations in our materials appeared verbatim in the 90M word training corpus; for those combinations, the model fell back on unigram probabilities, which in this context amounts to selecting the more frequent form of the verb.

Both RNNs performed much better than the \mbox{\emph{n}-gram} model on the simple agreement case (single-task: 94\%; multi-task: 100\%), reflecting these models' ability to generalize beyond the specific bigrams that occurred in the corpus. Accuracy on agreement in a sentential complement was also very high (single-task: 99\%; multi-task: 93\%). This indicates that the RNNs do not rely on the heuristic whereby the first noun of the sentence is likely to be its subject. They did slightly worse but still very well on short VP coordination (both 90\%); this dependency is also local, albeit across the word \textit{and}. 

\paragraph{Non-local agreement:} The accuracy of the \mbox{\emph{n}-gram model} on non-local dependencies (long VP coordination and agreement across a phrase or a clause) was very close to 50\%. This suggests that local collocational information is not useful in these conditions. The single-task RNN also performed much more poorly on these conditions than on the local agreement conditions, though for the most part its accuracy was better than chance. Humans did worse on these dependencies as well, but their accuracy did not drop as sharply as the RNNs' (human accuracies ranged from 82\% to 88\%). In most of these cases, multi-task learning was very helpful; for example, accuracy in long VP coordination increased from 61\% to 81\%. Still, both RNNs performed poorly on agreement across an object RC, especially without \textit{that}, whereas humans performed comparably on all non-local dependencies.

\paragraph{Agreement inside an object RC:} This case is particularly interesting, because this dependency is purely local (see \ref{ex:agr_in_orc}), and the interference is from the distant sentence-initial noun. Although this configuration is similar to the sentential complement case, performance was worse both in RNNs and humans. However, RNNs performed \textit{better} than humans, at least when the sentence included the overt relativizer \textit{that}. This suggests that interference is sensitive to proximity in RNNs but to syntactic status in humans --- humans appear to be confusing the main clause subject and the embedded subject \citep{wagers2009agreement}.

\paragraph{Reflexive anaphora:} The RNNs' performance was significantly worse on simple reflexives (83\%) than on simple agreement (94\%), and did not differ between the single-task and multi-task models. By contrast, human performance did not differ between subject-verb agreement and reflexive anaphoras. The surprisingly poor performance for this adjacent dependency seems to be due to an asymmetry in accuracy between \textit{himself} and \textit{themselves} on the one hand (100\% accuracy in the multi-task RNN) and \textit{herself} on the other hand (49\% accuracy).\footnote{This may be because \textit{himself} and \textit{themselves} are significantly more frequent than \textit{herself}, and consequently the number representation learned for \textit{herself} was not robust. Another possibility is that gender bias reduces the probability of an anaphoric relation between \textit{herself} and words such as \textit{surgeon} \cite{rudinger2018gender}.} Accuracy was very low for all pronouns in the structurally complex case in which the dependency was across a relative clause (55\% compared to 87\% in humans).

\paragraph{NPIs:} 

The dependency in simple NPIs spans only four words, so the \emph{n}-gram model could in principle capture it. In practice, the \emph{n}-gram model systematically selected the wrong answer, suggesting that it backed off to comparing the bigrams \textit{no students} and \textit{most students}, the first of which is presumably less frequent. Surprisingly, the \mbox{\emph{n}-gram} model's accuracy was higher than 50\% on NPIs across a relative clause, a dependency that spans more than five words. In this case, the bigrams \textit{that the} and \textit{the chef} (for example) happen to be more frequent than the \textit{that no} and \textit{no chef}. This difference was apparently strong enough to make up for the low-frequency bigram at the start of the sentence. 

The RNNs did poorly on this task. The accuracy of the single-task model was around 40\%. The multi-task did somewhat better on the simple NPIs (48\%) and much better on the NPIs across a relative clause (73\%). At the same time, an examination of the plot of log probability of each word in a sentence (Figure A.1 in the Supplementary Materials) suggests that the single-task RNN is in fact able to differentiate between the grammatical and ungrammatical sentences when it reaches the NPI, but this difference does not offset the overall probability advantage of the ungrammatical sentence (which is likely due to non-grammatical collocational factors). In any case, the fact that the \mbox{\emph{n}-gram} baseline did not perform at chance suggests that there are non-syntactic cues to this task, complicating the interpretation of the performance of other LMs. 

\paragraph{Perplexity:} The perplexity of the \mbox{\emph{n}-gram} model on the Wikipedia test data was 157.5, much higher than the perplexity of the single-task RNN (78.65) and the multi-task RNN (61.10). In other words, perplexity tracked accuracy on our syntactic data set \textendash\ an unsatisfying outcome given our goal of dissociating perplexity and our syntactic evaluation method, but an expected one given that each model was conditioned on richer information than the previous one. In previous work, perplexity and syntactic judgment accuracy have been found to be partly dissociable \cite{kuncoro2018lstms,tran2018importance}.

\begin{table*}
\centering
\scalebox{0.85}{
\begin{tabular}{ l l  c c c l }
\toprule
Main & Embedded & Single-task & Multi-task & Humans & Example sentence \\
subject & subject \\
\midrule
\multicolumn{3}{l}{\textit{Across an objective relative clause:}}\vspace{0.2em}\\
\underline{Singular} & Singular & 0.83 & 0.77 & 0.96 & The author that the minister likes laughs/*laugh. \\
\underline{Singular} & Plural & 0.51 & 0.30 & 0.90 & The author that the ministers like laughs/*laugh. \\
\underline{Plural} & Singular & 0.18 & 0.53 & 0.77 & The authors that the minister likes laugh/*laughs. \\
\underline{Plural} & Plural & 0.50 & 0.73 & 0.80 & The authors that the ministers like laugh/*laughs. \\
\vspace{0.2em}\\
\multicolumn{3}{l}{\textit{Within an objective relative clause:}}\vspace{0.2em}\\
Singular & \underline{Singular} & 0.73 & 0.92 & 0.94 & The author that the minister likes/*like laughs. \\
Singular & \underline{Plural} & 0.91 & 0.81 & 0.72 & The author that the ministers like/*likes laugh. \\
Plural & \underline{Singular} & 0.81 & 0.97 & 0.73 & The authors that the minister likes/*like laugh. \\
Plural & \underline{Plural} & 0.87 & 0.84 & 0.76 & The authors that the ministers like/*likes laugh. \\
\bottomrule
\end{tabular}
}
\caption{Accuracy within and across an object relative clause (only in the cases in which the main subject was animate and the relativizer \textit{that} was present). The subject that the verb is expected to agree with is underlined.}
\label{table:orc}
\end{table*}

\paragraph{Lexical variation and frequency:} There was considerable lexical variation in the results; we have mentioned the surprising asymmetry between \textit{himself} and \textit{herself} above. As another case study, we examine variation in the results of the simple agreement condition in the single-task RNN. Accuracy varied by verb, ranging from \textit{is} and \textit{are}, which had 100\% accuracy, to \textit{swims}, where accuracy was only 60\% (recall that average accuracy was 94\%). This may be a frequency effect: either the LM is learning less robust number representations for infrequent verbs, or the tail of the distribution over the vocabulary is more fragile during word prediction. \newcite{pauls2012large} propose normalizing for unigram frequency when deriving acceptability judgments from an LM. Our preliminary experiments with this method did not significantly improve overall performance; regardless of the effectiveness of this method, such corrections should arguably not be necessary in an LM that adequately captures grammaticality.

\section{Case study: agreement and object relative clauses}

The overall results in Table~\ref{table:overall} were averaged over all of the possible number configurations within each condition. In this section, we take a closer look at agreement in sentences with an object RC (see Table~\ref{table:orc}). This kind of finer-grained analysis helps explain the cases in which the LMs are failing, and might reveal some of the patterns or heuristics the LMs are using.

Performance in agreement across an object RC was poor. Both RNNs made attraction errors: they often preferred the verb that agreed in number with the irrelevant embedded subject to the verb that agreed with the correct main subject. The multi-task RNN showed greater symmetry between the simpler singular/singular and plural/plural cases, whereas the single-task RNN performed poorly even in these cases, often preferring a singular verb when both subjects were plural. This default preference for singular verbs matches the behavior of younger children \cite{franck2004normal}. 

Performance in agreement within an object RC was better; still, the single-task RNN  made the most errors when both subjects were singular, perhaps due to a heuristic in which a sentence with multiple subjects is likely to have a plural verb (as in coordination sentences). By contrast, the multi-task model seemed to have a general bias towards singular subjects in this condition. Incidentally, the human results with object RCs were also unexpected: while attraction errors when the two subjects differ in number are to be expected \cite{wagers2009agreement}, our participants made a sizable number of errors even when both subjects were plural. 

Despite the generally poor performance in object RCs, Figures A.2 and A.3 in the Supplementary Materials show that the single-task RNN is typically assigning a higher probability to the grammatical word of a minimal pair than to the ungrammatical word.

\section{Discussion}

We have described a template-based data set for targeted syntactic evaluation of language models. The data set consists of pairs of sentences that are matched except for their grammaticality; we consider a language model to capture the relevant aspects of the grammar of the language if it assigns a higher probability to the grammatical sentence than to the ungrammatical one.

An RNN language model performed very well on local subject-verb agreement dependencies, significantly outperforming an \emph{n}-gram baseline. This suggests that the task is a viable evaluation strategy. Even on simple cases, however, the RNN's accuracy was sensitive to the particular lexical items that occurred in the sentence; this would not be expected if its syntactic representations were fully abstract. The RNN's performance degraded markedly on non-local dependencies, approaching chance levels on agreement across an object relative clause. Multi-task training with a syntactic objective (CCG supertagging) mitigated this drop in performance for some but not all of the dependencies we tested. We conjecture that the benefits of the inductive bias conferred by multi-task learning will be amplified when the amount of training data is limited.

Our results contrast with the results of \citet{Gulordava18}, who obtained a prediction accuracy of 81\% on English sentences from their test corpus and 74\% on constructed sentences modeled after sentences from the corpus. It is likely that our sentences are more syntactically challenging than the ones they were able to find in the relatively small manually annotated treebank they used.

One limitation of our approach is that it is not always clear what constitutes a minimal grammaticality contrast. In the subject-verb agreement case, the contrast was clear: the two present-tense forms of the verb, e.g., \textit{laugh} vs. \textit{laughs}. Our NPI manipulations, on the other hand, were less successful: the members of the contrasts differed not only in their syntactic structure but also in low-level \emph{n}-gram probabilities, making the performance on this particular contrast harder to interpret.

We emphasize that the goal of this article was not to advocate for LSTMs in particular as an effective architecture for modeling syntax; indeed, our results show that LSTM language models are far from matching naive annotators' performance on this task, let alone performing at 100\% accuracy. We hope that our data set, and future extensions to other phenomena and languages, will make it possible to measure progress in syntactic language modeling and will lead to better understanding of the syntactic generalizations captured by language models. 

\section{Acknowledgments}
We would like to thank Ming Xiang for sharing materials from human experiments that inspired many of our test cases. We also thank Brian Roark and the JHU Computational Psycholinguistics lab for discussion, and Brian Leonard for help conducting the human experiment.

\bibliography{syntacticeval}
\bibliographystyle{acl_natbib}

\newpage
\appendix

\onecolumn

\section{Supplementary Materials}
\label{sec:supplemental}

\subsection{All of the lexical items}
\begin{enumerate}
\item \textit{Determiners:} the, some
\item \textit{Complementizers:} that
\item \textit{Other quantifiers:} most, many, no, few
\item \textit{Main subjects (animate):} author(s), pilot(s), surgeon(s), farmer(s), manager(s), customer(s), officer(s), teacher(s), senator(s), consultant(s)
\item \textit{Main subjects (inanimate):} movie(s), book(s), game(s), song(s), picture(s), painting(s), novel(s), poem(s), show(s)
\item \textit{Embedded subjects:} security guard(s), chef(s), architect(s), skater(s), dancer(s), minister(s), taxi driver(s), assistant(s), executive(s), parent(s)
\item \textit{Main verbs (animate):} laugh(s), swim(s), smile(s), is/are tall, is/are old, is/are young is/are short
\item \textit{Main verbs (inanimate):} is/are good, is/are bad, is/are new, is/are popular, is/are unpopular, bring(s) joy to people, interest(s) people
\item \textit{Long main verbs:} know(s) many different foreign languages, like(s) to watch television shows, is/are twenty three years old, enjoy(s) playing tennis with colleagues, write(s) in a journal every day
\item \textit{Reflexive main verbs:} hurt, injured, congratulated, embarrassed, disguised, hated, doubted
\item \textit{NPI verbs (animate):} be/been popular, be/been famous, have/had children
\item \textit{NPI verbs (inanimate):} be/been seen, be/been appreciated, be/been ignored, get/gotten old
\item \textit{Embedded verbs:} like(s), admire(s), hate(s), love(s)
\item \textit{Prepositions (animate):} next to, behind, in front of, near, to the side of, across from
\item \textit{Auxiliaries:} have, will
\item \textit{NPIs:} ever
\item \textit{Prepositions (inanimate):} from, by
\item \textit{Sentential complement subjects:} mechanic(s), banker(s)
\item \textit{Sentential complement verbs:} said, thought, knew
\item \textit{Conjunctions:} and
\item \textit{Anaphors:} himself, herself, themselves
\end{enumerate}

\subsection{All of the templates (single-task, multi-task)}

\begin{enumerate}
\item SUBJECT-VERB AGREEMENT
\begin{enumerate}
\item Simple:
\begin{enumerate}
\item Singular: \\
the author laughs/*laugh. : (0.93, 1.0)
\item Plural: \\
the authors laugh/*laughs. : (0.96, 1.0)
\end{enumerate}
\item In a sentential complement: 
\begin{enumerate}
\item Plural / \underline{Singular}:\\
the mechanics said the author laughs/*laugh. : (0.98, 0.86)
\item Singular / \underline{Singular}:\\
the mechanic said the author laughs/*laugh. : (0.98, 0.87)
\item Plural / \underline{Plural}:\\
the mechanics said the authors laugh/*laughs. : (1.0, 1.0)
\item Singular / \underline{Plural}:\\
the mechanic said the authors laugh/*laughs. : (1.0, 1.0)
\end{enumerate}
\item Short VP coordination:
\begin{enumerate}
\item Plural: \\
the authors laugh and swim/*swims. : (0.87, 0.91)
\item Singular: \\
the author laughs and swims/*swim. : (0.94, 0.88)
\end{enumerate}
\item Long VP coordination:
\begin{enumerate}
\item Singular: \\
the author knows many different foreign languages and enjoys/*enjoy playing tennis with colleagues. : (0.56, 0.82)
\item Plural: \\
the authors know many different foreign languages and enjoy/*enjoys playing tennis with colleagues. : (0.66, 0.80)
\end{enumerate}
\item Across a prepositional phrase:
\begin{enumerate}
\item Singular (animate) / Plural: \\
the author next to the security guards smiles/*smile. : (0.08, 0.37)
\item Singular (inanimate) / Plural: \\
the movie from the security guards smiles/*smile. : (0.12, 0.41)
\item Plural (animate) / Singular: \\
the authors next to the security guard smile/*smiles. : (0.33, 0.48)
\item Plural (inanimate) / Singular: \\
the movies from the security guard smile/*smiles. : (0.46, 0.57)
\item Singular (inanimate) / Singular: \\
the movie from the security guard smile/*smiles. : (0.84, 0.86)
\item Singular (animate) / Singular: \\
the author next to the security guard smile/*smiles. : (0.85, 0.91)
\item Plural (animate) / Plural: \\
the movies from the security guards smiles/*smile. : (0.98, 1.0)
\item Plural (inanimate) / Plural: \\
the authors next to the security guards smiles/*smile. : (1.0, 1.0)
\end{enumerate}
\item Across a subject relative clause:
\begin{enumerate}
\item \underline{Singular} / Plural: \\
the author that likes the security guards laughs/*laugh. : (0.05, 0.50)
\item \underline{Plural} / Singular: \\
the authors that like the security guard laugh/*laughs. : (0.30, 0.52)
\item \underline{Singular} / Singular: \\
the author that likes the security guard laughs/*laugh. : (0.89, 0.95)
\item \underline{Plural} / Plural: \\
the authors that like the security guards laugh/*laughs. : (1.0, 1.0)
\end{enumerate}
Across an object relative clause:
\begin{enumerate}
\item \underline{Plural (animate)} / Singular: \\
the authors that the security guard likes laugh/*laughs. : (0.18, 0.36)
\item \underline{Plural (inanimate)} / Singular: \\
the movies that the security guard likes are/*is good. : (0.25, 0.53)
\item \underline{Singular (inanimate)} / Plural: \\
the movie that the security guards like is/*are good. : (0.47, 37)
\item \underline{Plural (animate)} / Plural: \\
the authors that the security guards like laugh/*laughs. : (0.50, 0.73)
\item \underline{Singular (animate)} / Plural: \\
the author that the security guards like laughs/*laugh. : (0.51, 0.30)
\item \underline{Plural (inanimate)} / Plural: \\
the movies that the security guards like are/*is good. : (0.54, 0.74)
\item \underline{Singular (inanimate)} / Singular : \\
the movie that the security guard likes is/*are good. : (0.73, 0.78)
\item \underline{Singular (animate)} / Singular: \\
the author that the security guard likes laughs/*laugh. : (0.83, 0.77)
\end{enumerate}
\item Across an object relative (no \textit{that})
\begin{enumerate}
\item \underline{Plural (inanimate)} / Singular: \\
the movies the security guard likes are/*is good. : (0.30, 0.42)
\item \underline{Plural (animate)} / Singular: \\
the authors the security guard likes laugh/*laughs. : (0.33, 0.43)
\item \underline{Plural (animate)} / Plural: \\
the authors the security guards like laugh/*laughs. : (0.41, 0.52)
\item \underline{Plural (inanimate)} / Plural: \\
the movies the security guards like are/*is good. : (0.45, 0.50)
\item \underline{Singular (inanimate)} / Plural: \\
the movie the security guards like is/*are good. : (0.58, 0.59)
\item \underline{Singular (animate)} / Plural: \\
the author the security guards like laughs/*laugh. : (0.60, 0.54)
\item \underline{Singular (inanimate)} / Singular : \\
the movie the security guard likes is/*are good. : (0.71, 0.60)
\item \underline{Singular (animate)} / Singular: \\
the author the security guard likes laughs/*laugh. : (0.75, 0.60)
\end{enumerate}
\item In an object relative clause:
\begin{enumerate}
\item Singular (inanimate) / \underline{Singular} : \\
the movie that the security guard likes/*like is good. : (0.69, 0.88)
\item Singular (animate) / \underline{Singular}: \\
the author that the security guard likes/*like laughs. : (0.74, 0.92)
\item Plural (inanimate) / \underline{Singular}: \\
the movies that the security guard likes/*like are good. : (0.79, 0.94)
\item Plural (animate) / \underline{Singular}: \\
the authors that the security guard likes/*like laugh. : (0.81, 0.97)
\item Plural (animate) / \underline{Plural}: \\
the authors that the security guards like/*likes laugh. : (0.87, 0.84)
\item Singular (animate) / \underline{Plural}: \\
the author that the security guards like/*likes laughs. : (0.91, 0.81)
\item Plural (inanimate) / \underline{Plural}: \\
the movies that the security guards like/*likes are good. : (0.92, 0.88)
\item Singular (inanimate) / \underline{Plural}: \\
the movie that the security guards like/*likes is good. : (0.96, 0.90)
\end{enumerate}
\item In an object relative (no \textit{that})
\begin{enumerate}
\item Singular (inanimate) / \underline{Singular} : \\
the movie the security guard likes/*like is good. : (0.42, 0.65)
\item Plural (animate) / \underline{Singular}: \\
the authors the security guard likes/*like laugh. : (0.48, 0.89)
\item Singular (animate) / \underline{Singular}: \\
the author the security guard likes/*like laughs. : (0.50, 0.67)
\item Plural (inanimate) / \underline{Singular}: \\
the movies the security guard likes/*like are good. : (0.51, 0.89)
\item Plural (animate) / \underline{Plural}: \\
the authors the security guards like/*likes laugh. : (0.91, 0.86)
\item Singular (animate) / \underline{Plural}: \\
the author the security guards like/*likes laughs. : (0.94, 0.84)
\item Plural (inanimate) / \underline{Plural}: \\
the movies the security guards like/*likes are good. : (0.97, 0.88)
\item Singular (inanimate) / \underline{Plural}: \\
the movie the security guards like/*likes is good. : (0.98, 0.92)
\end{enumerate}
\end{enumerate}

\item REFLEXIVE ANAPHORA
\begin{enumerate}
\item Simple:
\begin{enumerate}
\item Singular: \\
the author injured himself/*themselves. : (0.67, 0.73)
\item Plural: \\
the authors injured themselves/*himself. : (0.99, 1.0)
\end{enumerate}
\item In a sentential complement:
\begin{enumerate}
\item Plural / \underline{Singular} \\
the mechanics said the author hurt himself/*themselves. : (0.71, 0.60)
\item Singular / \underline{Singular} \\
the mechanic said the author hurt himself/*themselves. : (0.80, 0.78)
\item Singular / \underline{Plural} \\
the mechanic said the authors hurt themselves/*himself. : (0.93, 0.96)
\item Plural / \underline{Plural} \\
the mechanics said the authors hurt themselves/*himself. : (1.0, 1.0)
\end{enumerate}
\item Across a relative clause:
\begin{enumerate}
\item \underline{Singular} / Plural: \\
the author that the security guards like injured himself/*themselves. : (0.10, 0.09)
\item \underline{Plural} / Singular: \\
the authors that the security guard likes injured themselves/*himself. : (0.33, 0.45)
\item \underline{Singular} / Singular: \\
the author that the security guard likes injured himself/*themselves. : (0.82, 0.75)
\item \underline{Plural} / Plural: \\
the authors that the security guards like injured themselves/*himself. : (0.93, 0.94)
\end{enumerate}
\end{enumerate}

\item NEGATIVE POLARITY ITEMS
\begin{enumerate}
\item Simple:
\begin{enumerate}
\item Past tense: \\
no/*most authors have ever been famous. : (0.35, 0.42)
\item Future tense: \\
no/*most authors will ever be famous. : (0.45, 0.55)
\end{enumerate}
\item Across a relative clause:
\begin{enumerate}
\item Past tense: \\
no authors that the security guards like have ever been famous. (grammatical) \\
vs. *the authors that no security guards like have ever been famous. (intrusive) : (0.37, 0.73) \\
\item Future tense: \\
no authors that the security guards like will ever be famous. (grammatical) : (0.44, 0.75) \\
\end{enumerate}
\end{enumerate}
\end{enumerate}

\subsection{Details of human experiment}

We constructed ten lists of minimal pairs. Each list had 76 minimal pairs: one representative of each of the 67 conditions, and nine additional Simple Agreement pairs included to detect participants who did not understand the task. Both versions of a minimal pair were shown on the screen at the same time; participants were asked to judge which one of them was more acceptable. The precise instructions were as follows:

\begin{quote}
In each trial, you will be presented with two sentences. You will read each sentence carefully and decide which of the two is more "acceptable" than the other. For instance, if the sentences were "The dogs in the park are happy" and "The dogs in the park is happy", then you would choose the first sentence, since it does not contain a grammatical error. You make click the sentence to choose it, or you may press the 1 or 2 key to choose. To ensure that your submission is accepted, be sure to read each sentence and choose your response carefully. Your goal should be accuracy, not speed.
\end{quote}

The order of the two members of the minimal pair was randomized: the grammatical sentence was shown above the ungrammatical one in half of the cases and below the ungrammatical one in the other half.

We recruited 100 participants on Amazon Mechanical Turk. Ten participants were assigned to each list, such that we collected ten judgments for each minimal pair. We excluded participants who made more than one error on the ten Simple Agreement sentences (16 participants).

\subsection{Word probabilities over the course of a sentence}

\begin{figure}
\includegraphics[width=\textwidth]{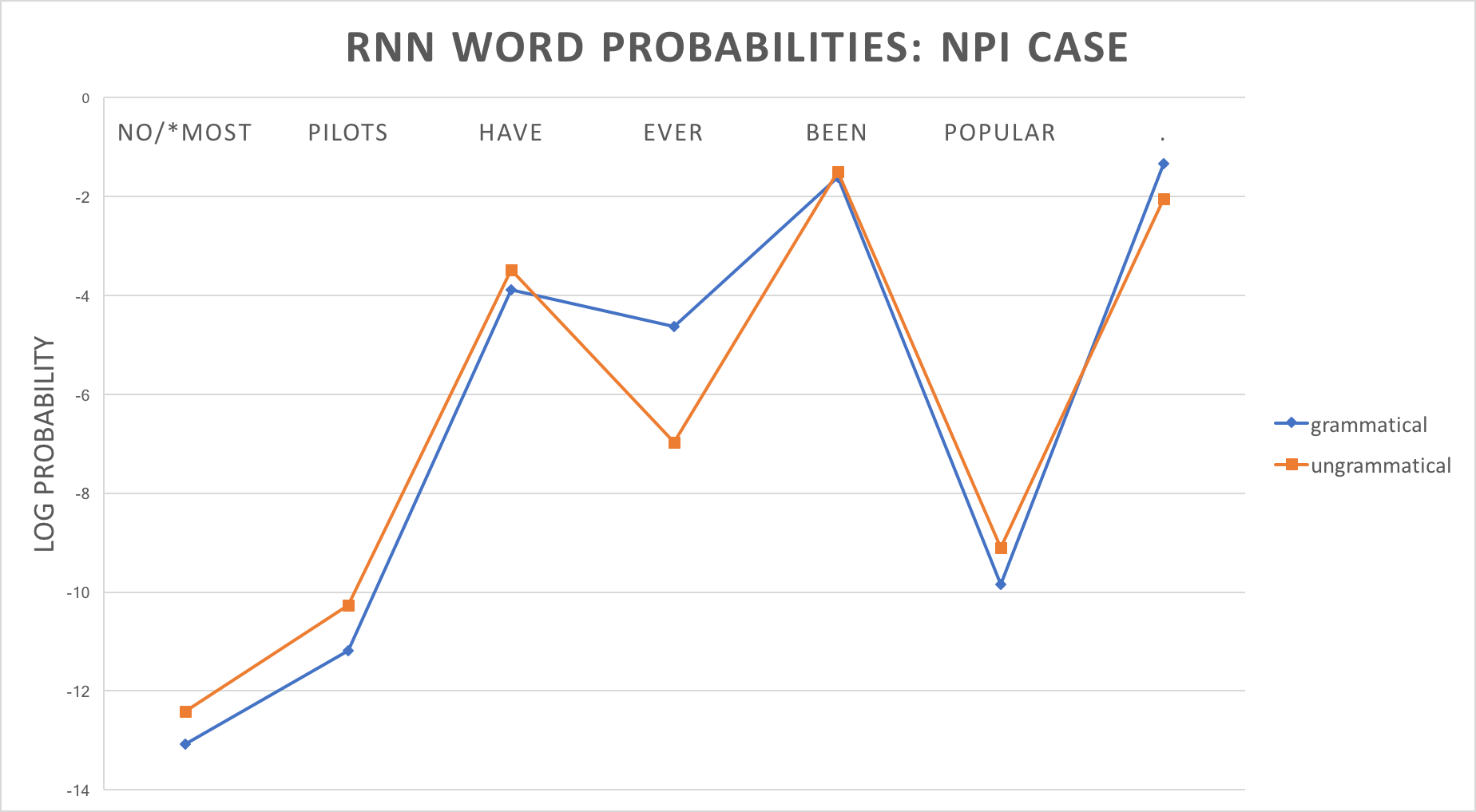}
\caption{The log probabilities for each word in the ``simple NPI'' case, as assigned by the single-task RNN.}
\label{fig:npi-wordprobs}
\end{figure}

\begin{figure}
\includegraphics[width=\textwidth]{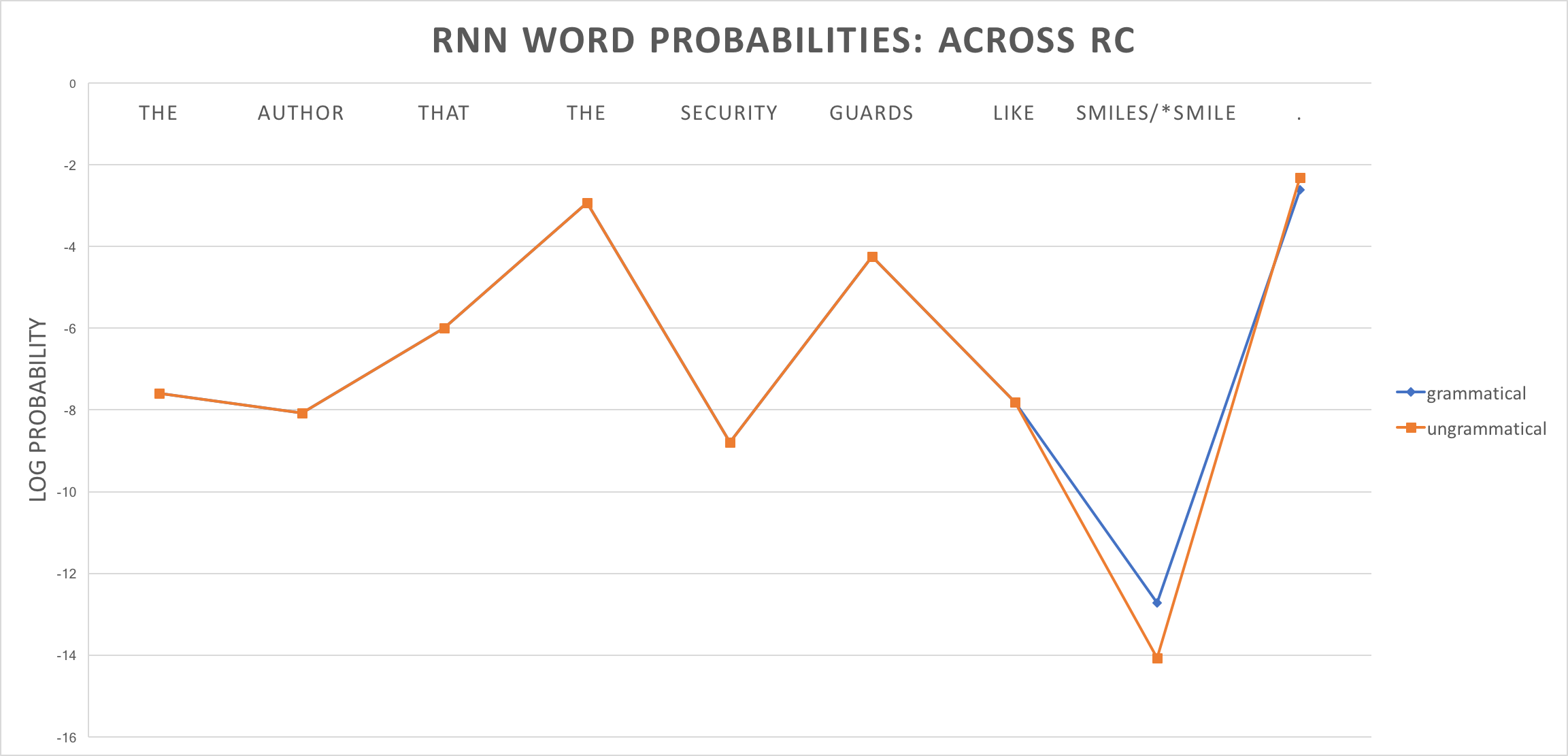}
\caption{The log probabilities for each word in the ``across an object relative clause'' case, as assigned by the single-task RNN.}
\label{fig:across-orc-wordprobs}
\end{figure}

\begin{figure}
\includegraphics[width=\textwidth]{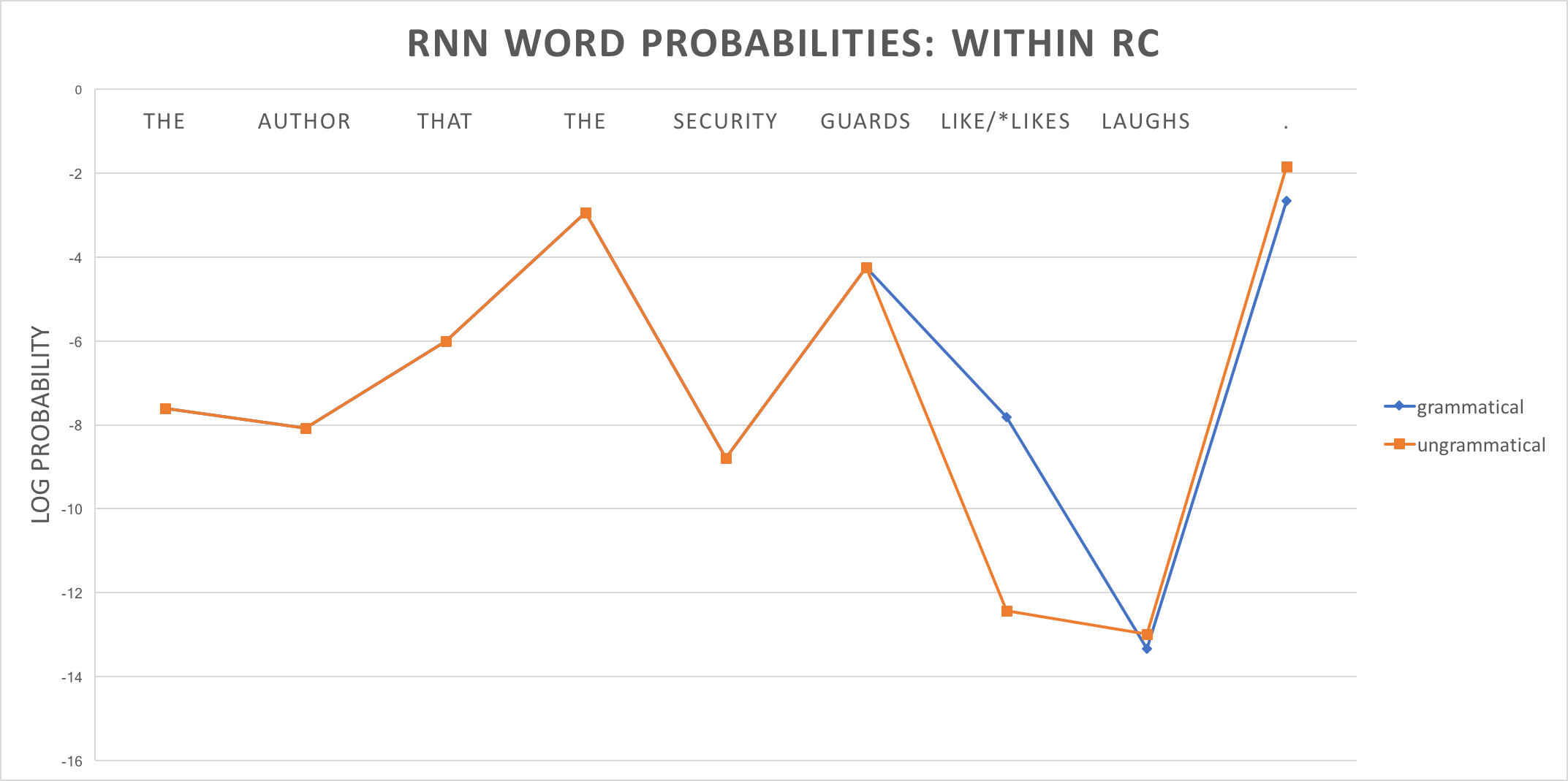}
\caption{The log probabilities for each word in the ``in an object relative clause'' case, as assigned by the single-task RNN.}
\label{fig:within-orc-wordprobs}
\end{figure}

\end{document}